\documentclass[runningheads]{llncs}
\usepackage{graphicx}

\usepackage{tikz}
\usepackage{comment}
\usepackage{amsmath,amssymb} 
\usepackage{color}

\usepackage[accsupp]{axessibility}  

\usepackage{empheq}
\usepackage{amsmath}
\usepackage{mathtools}

\usepackage{cases}
\usepackage{cite}
\usepackage{array}
\newcolumntype{L}[1]{>{\raggedright\let\newline\\\arraybackslash\hspace{0pt}}m{#1}}
\newcolumntype{C}[1]{>{\centering\let\newline\\\arraybackslash\hspace{0pt}}m{#1}}
\newcolumntype{R}[1]{>{\raggedleft\let\newline\\\arraybackslash\hspace{0pt}}m{#1}}

\usepackage{xcolor}
\usepackage{hyperref}
\hypersetup{
    colorlinks=true,
    citebordercolor=green,
    linkbordercolor=red,
    urlbordercolor=cyan,
}

\begin{document}
\pagestyle{headings}
\mainmatter
\def\ECCVSubNumber{6719}  

\title{Deforming Radiance Fields with Cages} 

\titlerunning{Deforming Radiance Fields with Cages}
%
\author{Tianhan Xu\inst{1} \and Tatsuya Harada\inst{1,2}}
\authorrunning{T. Xu and T. Harada}
%
\institute{The University of Tokyo \and
RIKEN \\\email{\{tianhan.xu, harada\}@mi.t.u-tokyo.ac.jp}}

\maketitle

\begin{abstract}
Recent advances in radiance fields enable photorealistic rendering of static or dynamic 3D scenes, but still do not support explicit deformation that is used for scene manipulation or animation. In this paper, we propose a method that enables a new type of deformation of the radiance field: free-form radiance field deformation. We use a triangular mesh that encloses the foreground object called \textit{cage} as an interface, and by manipulating the cage vertices, our approach enables the free-form deformation of the radiance field. The core of our approach is cage-based deformation which is commonly used in mesh deformation. We propose a novel formulation to extend it to the radiance field, which maps the position and the view direction of the sampling points from the deformed space to the canonical space, thus enabling the rendering of the deformed scene. The deformation results of the synthetic datasets and the real-world datasets demonstrate the effectiveness of our approach. Project page: \url{https://xth430.github.io/deforming-nerf/}.
\end{abstract}

\section{Introduction}

Photorealistic free-view rendering has recently received increasing attention for its various real-world applications such as virtual reality, augmented reality, games, and movies. Recently, neural scene representations~\cite{sitzmann2019scene, niemeyer2020differentiable, liu2020neural, mildenhall2020nerf} have shown better capability to capture both geometry and appearance that exceed traditional structure-from-motion~\cite{triggs1999bundle, hartley2003multiple} or image-based rendering~\cite{gortler1996lumigraph, davis2012unstructured}. The most representative work is Neural Radiance Field (NeRF)~\cite{mildenhall2020nerf}, which represents the static 3D scene as a radiance field and uses a neural network to encode the volume density and the view-dependent radiance color. With volume rendering~\cite{kajiya1984ray}, NeRF can achieve photorealistic rendering from an arbitrary viewpoint. Subsequent works extended NeRF to support modeling dynamic scenes~\cite{pumarola2021d, park2021nerfies, park2021hypernerf, tretschk2021non}, dark scenes~\cite{mildenhall2022nerf}, multi-scale rendering~\cite{barron2021mip}. Manipulable or editable scene rendering is one of the directions of NeRF extensions that received attention for its numerous applications such as scene animation or new scene generation. However, the above-mentioned works focus on modeling the existing scenes and thus cannot generate scenes that are unseen during the training.

For some specific object categories, such as the human body or articulated objects, recent studies~\cite{peng2021neural, peng2021animatable, liu2021neural, xu2022surface, 2021narf, su2021anerf, noguchi2022watch} enable the generation of the unseen scene by controlling the body shape or bone pose. Besides, some works utilize the idea of compositionality to separate foreground objects in the scene during training, thus allowing the scaling or moving of objects in the scene~\cite{yang2021objectnerf, zhang2021stnerf}. However, the manipulation in these approaches only allows affine transformations of objects. Although the above methods attempted to develop for radiance field manipulation, they have a common and clear limitation: they cannot perform explicit scene manipulation with details (e.g. torsion or local scaling) for arbitrary categories of objects.

To address the above issues, we propose a new approach for manipulating the optimized radiance field. Our method allows free-form deformation of the radiance field, thus enabling explicit object-level scene deformation or animation. Our idea is an extension of cage-based deformation (CBD), which is originally proposed for mesh deformation~\cite{ju2005mean, joshi2007harmonic, lipman2008green}. Specifically, the deformation of a fine mesh, or the displacement of its vertices, is driven by manipulating the vertices of the coarse triangular mesh called \textit{cage} that enclosed the fine mesh (e.g. Fig.~\ref{fig:cage-generation}(c)). Such a mesh deformation method is also known as \textit{free-form deformation}. Extending cage-based deformation to radiance field deformation while maintaining the properties of the radiance field such as volumetric representation and view-dependent radiance is non-trivial and yet unexplored. In this paper, we derive a novel formulation for applying CBD to the radiance field that satisfies the properties of the radiance field. However, we find that simply applying the proposed formulation to achieve radiance field deformation brings a new issue: the volume rendering process of the radiance field usually requires a huge number of sampling points~\cite{mildenhall2020nerf}, and CBD is usually accompanied by a high-dimensional tensor computation, these facts lead to impractical deformation computation times. To address this specific issue, we also propose a discretization method specifically suitable for the radiance field that significantly reduces the computation time of CBD.

We conducted extensive experiments with various types of CBD algorithms using synthetic datasets and real-world datasets. Reasonable deformation and photorealistic rendering quality demonstrate the effectiveness of our approach.

\begin{figure}[t]
\centering
\includegraphics[width=\linewidth]{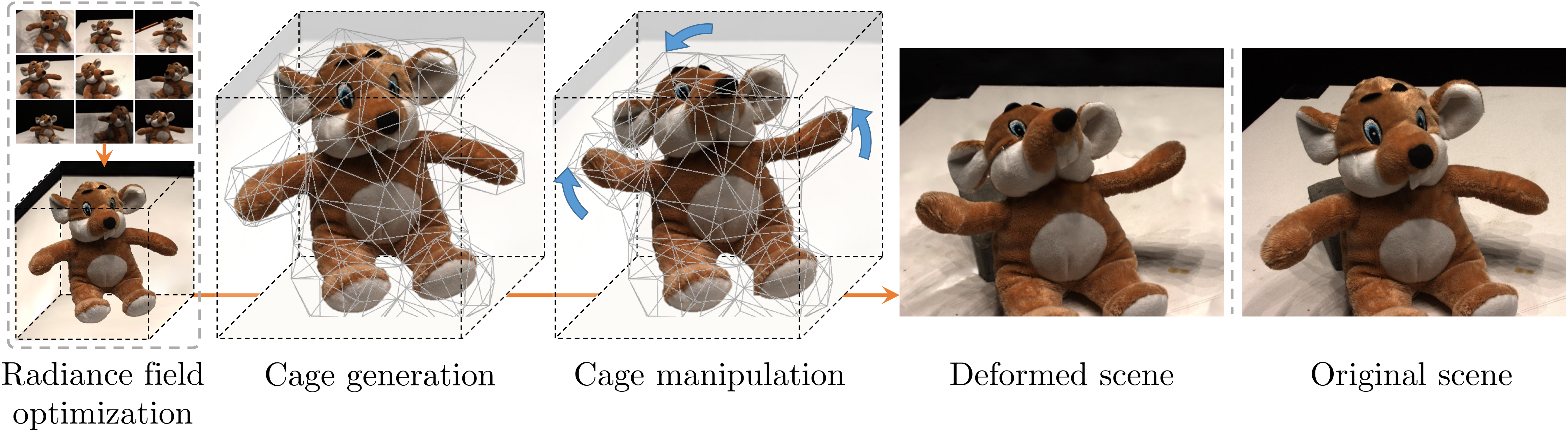}
\caption{An overview of our approach. Our method takes multi-view images capturing a static scene as input and uses an off-the-shelf algorithm to optimize a radiance field. Then, we automatically and/or manually generate a cage based on the optimized radiance field. By manipulating the vertices of the cage, the radiance field can be deformed accordingly. Finally, through volume rendering, the free-view rendering of the deformed scene can be achieved.}
\label{fig:teaser}
\end{figure}

In summary, our contributions are listed as follows:
\begin{itemize}
    \item We proposed a new approach to explicitly manipulate the radiance fields using a coarse triangular mesh called cage, allowing free-form deformation of the scene while maintaining photorealistic rendering quality.
    \item We proposed a discretization method for cage coordinate computation specifically adapted for the radiance field rendering, which achieves a speedup of several orders of magnitude compared to the naive computation.
    \item We conducted extensive experiments to deform the radiance field and the rendering results demonstrate the soundness and effectiveness of our method.
\end{itemize}

\section{Related Work} \label{sec:related}

\paragraph{\textbf{Neural scene representation.}}
Recently, neural scene representation, which uses a neural network to encode the 3D scenes, has received a lot of attention due to its high quality of geometry and appearance modeling compared to standard 3D representation including voxel~\cite{yan2016perspective, girdhar2016learning}, point clouds~\cite{fan2017point, guo2021pct} or textured-mesh~\cite{kanazawa2018learning, kato2018neural}. The most representative work is Neural Radiance Field (NeRF)~\cite{mildenhall2020nerf}, which shows that representing static scenes with volumetric density and view-dependent radiance can capture high-resolution geometry and support photorealistic novel view rendering. An obvious limitation of the original NeRF is that it can only model static scenes. Subsequent work relaxed this limitation and enabled the dynamic scene modeling by simultaneously learning the deformation fields~\cite{pumarola2021d, park2021nerfies, tretschk2021non} or introducing high-dimensional representation~\cite{park2021hypernerf}. While these methods achieved the capture of dynamic scenes, none of them can generate new dynamics that are unseen in the training.

\paragraph{\textbf{Manipulable neural scene rendering.}}
Recent work attempted to incorporate controllability into NeRF to achieve scene manipulation or new scene generation. For the specific task of human body modeling, various works proposed to combine NeRF with a parametric human model to enable human body re-posing~\cite{peng2021neural, peng2021animatable}, shape control~\cite{liu2021neural} or even clothing changes~\cite{xu2022surface}. For the articulated object, \cite{2021narf, su2021anerf} proposed to build NeRF on the local coordinates of the pre-defined skeleton thus allowing the rendering of the re-posed object, and \cite{noguchi2022watch} proposed to learn the unknown skeleton structure along with NeRF. However, the above approaches are limited to specific categories of objects and thus cannot be generalized to the modeling and manipulating of arbitrary objects.

In addition to the above methods of using human model or skeleton to assist in modeling, another direction of manipulatable scene modeling methods utilize an idea of compositionality~\cite{guo2020osf, Ost_2021_CVPR, yang2021objectnerf, zhang2021stnerf}. Specifically, these methods treat the 3D scene as a composition of multiple objects or backgrounds. By modeling each object independently, the movement or scaling of each object can be achieved. However, the controllability of such methods focuses on the location or size of objects w.r.t. the whole scene, and cannot achieve detailed deformation of the shape or appearance for individual objects. In contrast to all the above approaches, our method focuses on object-level deformation for detailed shape and appearance manipulation.

Concurrent work~\cite{yuan2022nerf-editing} uses a similar idea of mesh-based deformation for geometry editing of NeRF, which takes extracted fine mesh as an interface.

\paragraph{\textbf{Cage-based deformation.}}
Cage-based deformation (CBD) is a volumetric deformation method that is typically used for fine mesh deformation by manipulating the corresponding cage vertices. Here, \textit{cage} denotes a watertight mesh that encloses the target fine mesh to be deformed. The core of CBD is \textit{cage coordinates}, a generalized form of barycentric coordinates, which is used to represent the relative positions of spatial points w.r.t. the cage. The new position of a spatial point can be computed from its cage coordinates and the deformed cage. Previous works proposed several cage coordinates with different properties, including mean value coordinates (MVC)~\cite{floater2003mean, ju2005mean}, harmonic coordinates (HC)~\cite{derose2006harmonic, joshi2007harmonic}, green coordinates (GC)~\cite{lipman2008green}, etc. For example, the computation of MVC and GC have closed-formulation and thus can be computed in a feedforward manner, while HC does not have a closed-formulation and therefore its computation requires loop optimization. Specifically, the computation of HC discretizes the space into grid points and updates the HC value for each grid point by performing laplacian smoothing with certain boundary conditions. More comparisons and mathematical preliminaries can be found in~\cite{nieto2013cage}. In addition to the traditional CBD algorithm, recent works proposed to combine CBD with deep learning algorithm to achieve high-quality mesh deformation~\cite{Yifan:NeuralCage:2020, jakab2021keypointdeformer}. 

All of the above methods are focused on using CBD for mesh deformation. Our method aims to extend the CBD to the deformation of the radiance field.

\section{Method}

Our goal is to deform the optimized radiance field by manipulating the corresponding cage vertices, thus achieving a photorealistic rendering of the new deformed scene. An overview of our approach is shown in Fig.~\ref{fig:teaser}. The first step is to optimize a radiance field from the multi-view images~(Sec.~\ref{sec:radiance-field}). Then, a cage enclosing the foreground object is generated based on the optimized radiance field~(Sec.~\ref{sec:cage-generation}). With our proposed cage-based deformation formulation for the radiance field, the free-form deformation of the radiance field can be achieved~(Sec.~\ref{sec:cage-based-deformation}, Sec.~\ref{sec:deforming-radiance-field}). 

\begin{figure}[t]
\centering
\includegraphics[width=\linewidth]{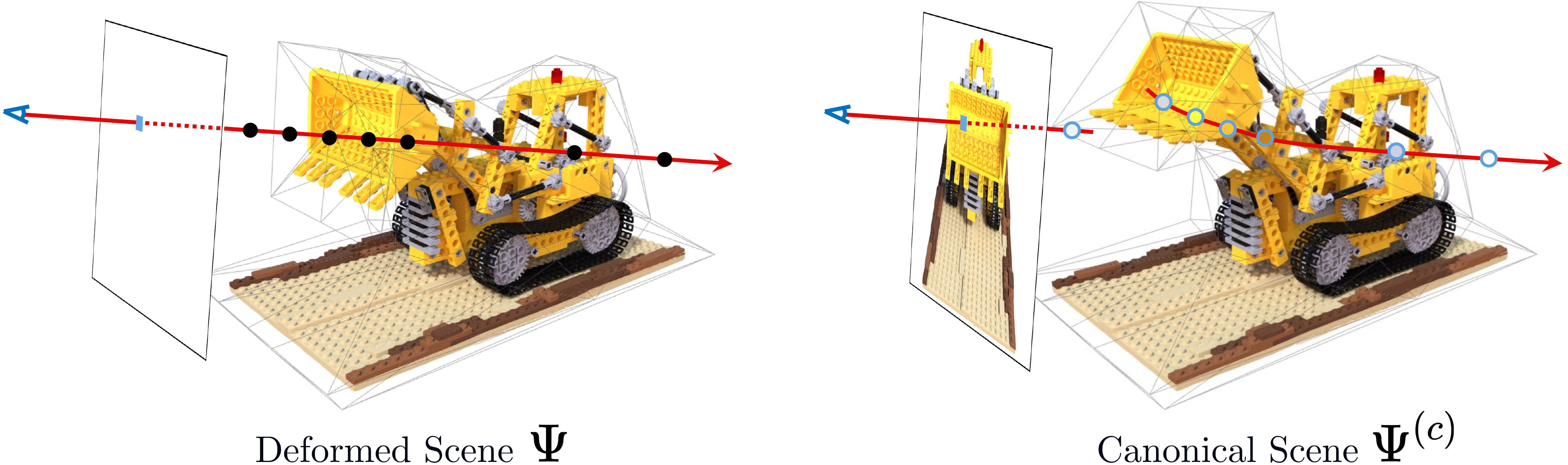}
\caption{Rendering process of the deformed scene. To perform volume rendering for the deformed radiance field $\mathrm{\Psi}$, we map the sampling points on the ray to the canonical space through cage-based deformation and query the color and density in the canonical radiance field $\mathrm{\Psi}^{(c)}$.}
\label{fig:pipeline}
\end{figure}

\subsection{Radiance fields revisited} \label{sec:radiance-field}
Neural radiance field (NeRF)~\cite{mildenhall2020nerf} uses a neural network to encode the 3D scene as a continuous neural representation, which receives the spatial position $\mathbf{x} \in \mathbb{R}^3$ and view direction $\mathbf{d} \in \mathbb{R}^3$ as inputs and computes the RGB color $\mathbf{c} \in \mathbb{R}^3$ and density $\sigma \in \mathbb{R}$ of that point. With volume rendering, photorealistic rendering of NeRF from an arbitrary viewpoint can be achieved. Recently, some variants of radiance field representation have been proposed, for example, Plenoxels~\cite{yu_and_fridovichkeil2021plenoxels} use grid representation and directly optimize radiance field without using neural networks. Without loss of generality, we refer to the 3D scene representation that can be formulated as $\mathrm{\Psi}: (\mathbf{x}, \mathbf{d}) \rightarrow (\mathbf{c}, \sigma)$ as \textit{radiance field}.

As explored in previous studies, given a static scene, a radiance field $\mathrm{\Psi}^{(c)}$ can be optimized from a set of multi-view images with calibrated camera parameters. Here $c$ stands for ``canonical'', which denotes the original static scene, to distinguish it from the later deformed scene.

\subsection{Cage generation from radiance fields} \label{sec:cage-generation}
\begin{figure}[t]
\centering
\includegraphics[width=\linewidth]{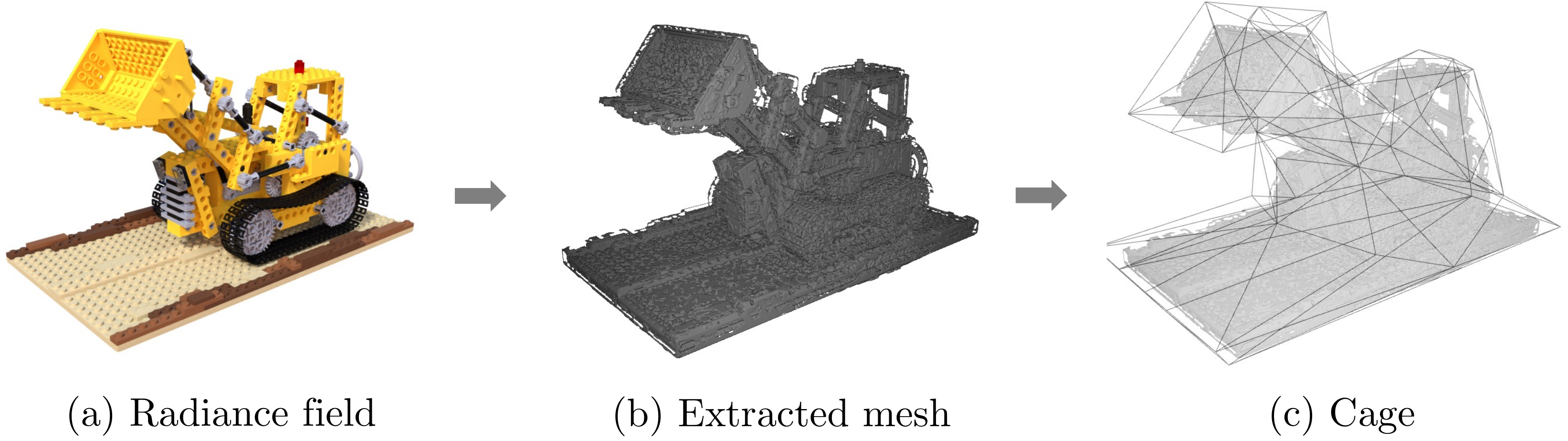}
\caption{Cage generation process. After extracting a fine mesh from the optimized radiance field, a cage can be generated automatically and/or manually according to the fine mesh.}
\label{fig:cage-generation}
\end{figure}

In this paper, \textit{cage} refers to a coarse 3D triangular mesh that strictly encloses the foreground object. We demonstrate a method for automatically and/or manually generating a cage from the optimized radiance field. Specifically, the first step is to convert the radiance field into a fine mesh using surface extraction methods such as marching cubes~\cite{lorensen1987marching} (Fig.~\ref{fig:cage-generation}(b)). The second step is to create the corresponding cage for the generated mesh (Fig.~\ref{fig:cage-generation}(c)). For scenes containing only foreground objects (such as those optimized using masked images), we use~\cite{xian2009automatic} to compute the corresponding cage. For scenes containing backgrounds, we use Blender~\cite{blender} to manually split the foreground objects from the reconstructed fine mesh and then apply~\cite{xian2009automatic} to compute the cage. However, some cage predictions may be inaccurate due to the complex shapes or fine details of the scenes. For these cases, we manually resolve them based on the automatically generated cage for better manipulation performance. Alternatively, if only a simple deformation (or moving, scaling) is needed, we can also use 3D software to manually build a simple cage, such as rectangle or cylinder, by referring to the extracted fine mesh.

\subsection{Cage-based deformation} \label{sec:cage-based-deformation}
Cage-based deformation (CBD) is originally proposed for deforming a fine mesh using the cage, which calculates the vertex displacement of the fine mesh caused by the cage manipulation. Specifically, given a cage $\mathcal{C}$ with vertices $\{\mathbf{v}_j\}$, points $\mathbf{x} \in \mathbb{R}^3$ inside $\mathcal{C}$ can be identified with cage coordinates $\{ \omega_j \}$ which represent the relative position of $\mathbf{x}$ w.r.t. $\mathcal{C}$. Formally, the position of point $\mathbf{x}$ is weighted by the cage vertices as:
\begin{align}
    \mathbf{x} = \sum_j \omega_j(\mathbf{x}) \mathbf{v}_j .
\end{align}
Consider that we manipulate the vertices of $\mathcal{C}$ and deform it to cage $\mathcal{C}'$ with vertices $\{\mathbf{v}'_j\}$. Using the calculated cage coordinate, the deformed position of $\mathbf{x}$ for the deformed cage $\mathcal{C}'$ can be calculated as:
\begin{align} \label{equ:cbd2}
    \mathbf{x}' = \sum_j \omega_j(\mathbf{x}) \mathbf{v}'_j .
\end{align}
Previous studies~\cite{ju2005mean, joshi2007harmonic, lipman2008green} proposed several kinds of cage coordinates and achieved promising results on the mesh deformation.

Note that although the above formulation seems simple, the actual derivation and computation of the cage coordinate $\{ \omega_j \}$ is complicated and usually accompanied by a large tensor computation. For detailed computation, we recommend referring to the original papers of these cage coordinates~\cite{ju2005mean, joshi2007harmonic, lipman2008green}.

\subsection{Deforming radiance fields} \label{sec:deforming-radiance-field}
In this section, we introduce a novel formulation that extends the application of CBD from mesh to the radiance field. Remind that our goal is to deform the optimized radiance field $\mathrm{\Psi}^{(c)}$ for the free-view rendering of the deformed scenes. Suppose we have a cage $\mathcal{C}^{(c)}$ that accompanies $\mathrm{\Psi}^{(c)}$ that encloses the foreground object. Consider that we manipulate the vertices of $\mathcal{C}^{(c)}$ and deform it to a new cage $\mathcal{C}$, and denote the desired radiance field after deformation as $\mathrm{\Psi}$.

To achieve volume rendering of $\mathrm{\Psi}$, the sampling points are required to be mapped from the deformed space to the canonical space for color and density computations, as shown in Fig~\ref{fig:pipeline}. To describe such deformed-to-canonical mapping, we reversely treat the cage deformation process as: the new cage $\mathcal{C}$ is deformed to the canonical cage $\mathcal{C}^{(c)}$. While contrary to the actual cage manipulation process, such a convention allows us to map the points in the deformed space back to the canonical space. Specifically, we denote the deformed-to-canonical mapping of spatial position and view direction as:
\begin{align} \label{eq:d2c}
    \phi_{\mathbf{x}}: \mathbf{x} \rightarrow \mathbf{x}^{(c)}, \quad \phi_{\mathbf{d}}: (\mathbf{x}, \mathbf{d}) \rightarrow \mathbf{d}^{(c)}
\end{align}
Here, $\mathbf{x}^{(c)}$ can be simply derived from Eq.~(\ref{equ:cbd2}) and $\mathbf{d}^{(c)}$ can be derived from difference approximation as $\mathbf{d}^{(c)} = \mathrm{norm}( (\phi_{\mathbf{x}}(\mathbf{x} + \mathrm{\Delta}t \mathbf{d}) - \phi_{\mathbf{x}}(\mathbf{x})) / \mathrm{\Delta}t )$. $\mathrm{\Delta}t$ denotes a small constant and $\mathrm{norm}(\cdot)$ normalizes the vector length to $1$. Note that the above mappings are derived from the simple CBD computation without any learnable components.

The deformed radiance field $\mathrm{\Psi}$ can be divided into three parts depending on the space that: (1) outside both the canonical cage and deformed cage (2) inside the canonical cages but outside the deformed cages, (3) inside the deformed cages. Specifically, it can be formulated as follows:


\begin{subnumcases}{\label{eq:allcase} \mathrm{\Psi}(\mathbf{x},\mathbf{d}) = }
    \mathrm{\Psi}^{(c)}(\mathbf{x}, \mathbf{d}), & $\mathbf{x} \in \mathbb{R}^3 \setminus (\mathbb{V}^{(c)} \cup \mathbb{V})$ \label{eq:case1} \\
    (\mathbf{0}, 0), & $\mathbf{x} \in \mathbb{V}^{(c)} \setminus (\mathbb{V}^{(c)} \cap \mathbb{V})$ \label{eq:case2} \\
    \mathrm{\Psi}^{(c)} \left( \phi_{\mathbf{x}}(\mathbf{x}), \phi_{\mathbf{d}}(\mathbf{x}, \mathbf{d}) \right), & $\mathbf{x} \in \mathbb{V}$ \label{eq:case3}
\end{subnumcases}

Here, $\mathbb{V}^{(c)}, \mathbb{V} \subset \mathbb{R}^3$ denotes the space enclosed by $\mathcal{C}^{(c)}$ and $\mathcal{C}$, respectively. Eq.~(\ref{eq:case1}) indicates that the radiance field remains unchanged before and after deformation for the position outside the cages. For the points inside the canonical cage, we clear them, namely setting the color and density to zero, as in Eq.~(\ref{eq:case2}). For the points inside the deformed cage, we map the spatial position and view direction to the canonical space through Eq.~(\ref{eq:d2c}) and then query the color and density from $\mathrm{\Psi}^{(c)}$, as in Eq.~(\ref{eq:case3}).

\section{Implementation details}

\subsection{Faster cage coordinates computation} \label{sec:fast}
Technically, the rendering of the deformed scene can be achieved by computing the deformed-to-canonical mapping in Eq.~(\ref{eq:allcase}) for all the sampling points on all the rays. However, because of the huge number of sampling points and the fact that the computation of cage coordinates is usually accompanied by a high-dimensional tensor computation (as discussed in Sec.~\ref{sec:cage-based-deformation}), the above brute-force computation is usually impractical either in terms of time or memory capacity. A rough estimation can be given as, rendering images of size $(h, w)$ from $N$ different viewpoints, with $M$ points sampled on each ray, the number of points that require the cage coordinate computation is about $h \times w \times N \times M$ in order of magnitude\footnote[1]{In fact, the actual number is smaller than this approximation since we only compute for points inside the cage.}. For instance, for $M=512$, rendering $200$ images of size $(800, 800)$ requires about $\sim 10^{10}$ orders of magnitude in the times of cage coordinates computation.

Inspired by the computation process of harmonic coordinates (HC), we propose to discretize the space into $n\times n \times n$ grid points for cage coordinates computation, even for the cage coordinates that have their closed-formulations, i.e. MVC and GC (briefly discussed in Sec.~\ref{sec:related}). At the inference time, we pre-compute the cage coordinates for each grid and use trilinear interpolation to calculate the cage coordinates for arbitrary points. We surprisingly find that such a simple discretization, however, brings great benefits for the specific nature of volume rendering of the radiance field. Note that such discretization makes the number of cage coordinates computation independent of $h, w, N, M$ given above, that is, once the pre-computation of grid points is completed, there is no increase in the computation of cage coordinates when we want to render the scene from the additional new viewpoint or with different image resolution. The only thing to consider here is the size of $n$, which requires a trade-off between discretization resolution and computation speed. Here, the number of points that require cage coordinates computation is about $n^3$ in order of magnitude\footnotemark[1].

We practically use $n=128$ in our experiments which gives about $\sim 10^6$ orders of magnitude in the times of computation. 

\subsection{Cage refinement}
The computation complexity of cage coordinates is proportional to the number of cage vertices. For fast inference, we control the hyperparameters (e.g. discrete voxel size) in cage generation algorithm~\cite{xian2009automatic} to ensure that the number of vertices is in the range of $30 \sim 200$. For scenes with complex shapes or details, we first generate a cage with a larger number of vertices ($\sim 1000$) and then manually decimate the vertices using Blender~\cite{blender}.

\subsection{Radiance fields representation}
We use Plenoxels~\cite{yu_and_fridovichkeil2021plenoxels} as radiance fields representation, which supports very fast scene optimization and rendering. Note that our method is not dependent on specific radiance field representations and thus can be directly applied to other representations such as NeRF~\cite{mildenhall2020nerf} or the latest faster radiance field representations~\cite{SunSC22, mueller2022instant, tensorf}.

\section{Experiments}
In this section, we evaluate the effectiveness of our approach through a variety of scenes, including synthetic dataset and real-world dataset. We show the results of extensive ablation studies and then discuss the limitations of our approach.

Unless otherwise specified, the deformations of the results are performed with harmonic coordinate with discretization resolution $n=128$. For the canonical scene optimization, we follow the default setting used in \cite{yu_and_fridovichkeil2021plenoxels}. We use one Nvidia A100 GPU for all the experiments.

\subsection{Datasets}
\paragraph{\textbf{NeRF and NSVF synthetic dataset.}}
We use synthetic dataset in original NeRF~\cite{mildenhall2020nerf} and Neural Sparse Voxel Fields (NSVF)~\cite{liu2020neural} papers. These scenes contain only foreground objects, and the images are captured from multiple cameras placed on the hemisphere. We follow the train/test split as in the original papers.

\paragraph{\textbf{DTU MVS dataset.}}
We use the real-world DTU MVS dataset~\cite{jensen2014large}, which contains a variety of static objects, and each scene uses 49 or 64 cameras to capture high-resolution images. 
We use all available cameras for training, and create test camera trajectories from camera interpolation for evaluation.

\subsection{Results}
Since our approach is the first to use a coarse cage as an interface for free-form deformation of the radiance field, there is no existing method for a direct comparison. The ground truth of the deformed scene is also not available, therefore, we show the qualitative results before and after the deformation for evaluation.

\begin{figure}[t]
\centering
\includegraphics[width=\linewidth]{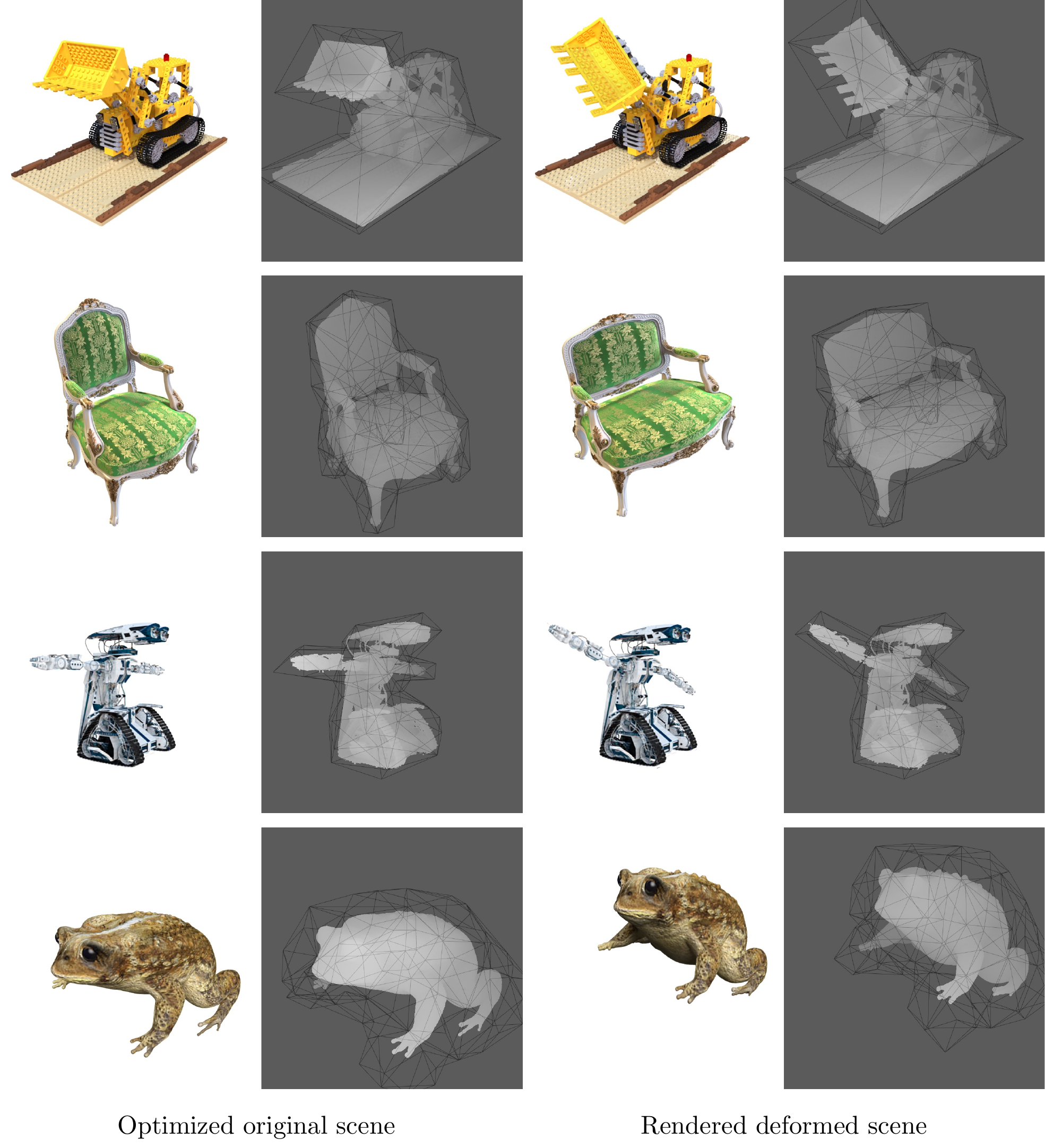}
\caption{Qualitative results on NeRF and NSVF synthetic dataset.
The original scene (left) and the deformed scene (right) are rendered from a novel viewpoint. Disparity map and corresponding cage are also presented.}
\label{fig:result_syn}
\end{figure}

\begin{figure}[t]
\centering
\includegraphics[width=\linewidth]{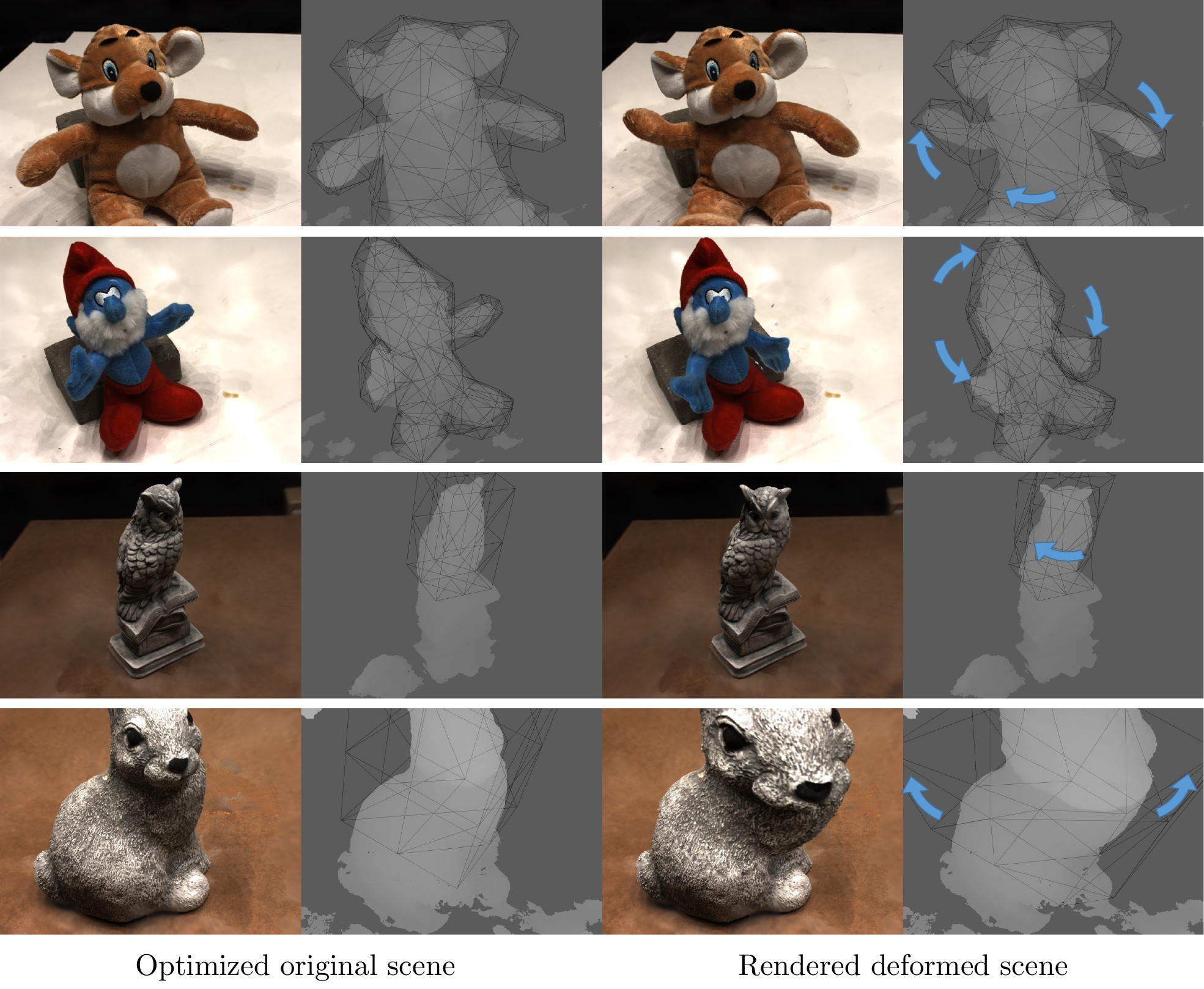}
\caption{Qualitative results on DTU dataset. The original scene (left) and the deformed scene (right) are rendered from a novel viewpoint. Disparity map of the foreground object and corresponding cage are also presented. Arrows illustrate the manipulation of the cage.}
\label{fig:result_dtu}
\end{figure}

We use Blender~\cite{blender} to manually deform the generated cage of the canonical scene with various types of deformations such as bending, stretching, torsion, scaling, etc. Novel view synthesis results of original/deformed scene on synthetic dataset and DTU dataset are shown in Fig~\ref{fig:result_syn} and Fig~\ref{fig:result_dtu}, respectively. As shown in the results, our proposed radiance field deformation approach enables explicit manipulation of the scene while maintaining photorealistic rendering quality. In addition to the free-form deformation of the entire object with the generated cage, our approach also allows for local manipulation of the object as in the last two figures in Fig.~\ref{fig:result_dtu}: all you need is to create a simple cage and deform it, which can be done with little effort using almost any existing 3D software.

\begin{figure}[t]
\centering
\includegraphics[width=\linewidth]{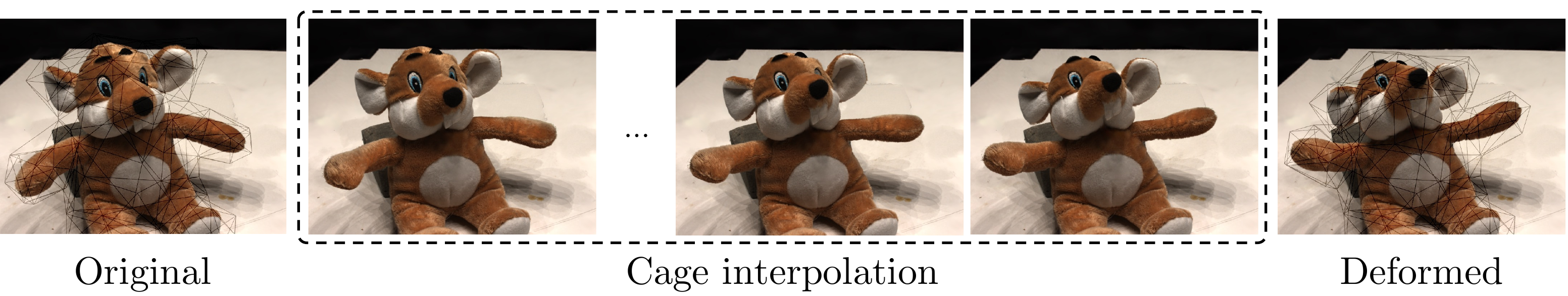}
\caption{Qualitative results of cage interpolation. Our approach can generate continuous free-view animation by interpolating the starting and ending cages.}
\label{fig:interp}
\end{figure}

The above features of our approach also support simple radiance field manipulation achieved by existing works~\cite{liu2021editing}, such as object movement, duplication, and scaling. Moreover, as shown in Fig.~\ref{fig:interp}, our method also supports applications of generating continuous free-view animation from a static scene by cage interpolation.

\begin{table}[t]
\begin{center}
\caption{
Computation time in seconds for rendering an image for three cage coordinates with different discretization resolutions. ``Precise'' means not using discretization, i.e., computing precise cage coordinates for all the sampling points on the rays. Here, ``MVC'' means mean value coordinates~\cite{ju2005mean}, ``HC'' means harmonic coordinates~\cite{joshi2007harmonic}, and ``GC'' means green coordinates~\cite{lipman2008green}. Please also refer to Sec.~\ref{sec:ablation} and Fig.~\ref{fig:abl_syn}.}
\label{table:abl}
\begin{tabular}{C{1.5cm}C{1.5cm}C{1.5cm}C{1.5cm}}
\hline\noalign{\smallskip}
 & MVC~\cite{ju2005mean} & HC~\cite{joshi2007harmonic} & GC~\cite{lipman2008green} \\
\hline
\noalign{\smallskip}
$64^3$  & 0.31 & 0.23 & 0.35 \\
$128^3$  & 0.98 & 0.90 & 2.49 \\
$256^3$  & 5.71 & 6.32 & 19.69 \\
\hline
\noalign{\smallskip}
Precise  & 102 & N/A & 243 \\
\hline
\end{tabular}
\end{center}
\end{table}
\setlength{\tabcolsep}{1.4pt}

\begin{figure}[t]
\centering
\includegraphics[width=\linewidth]{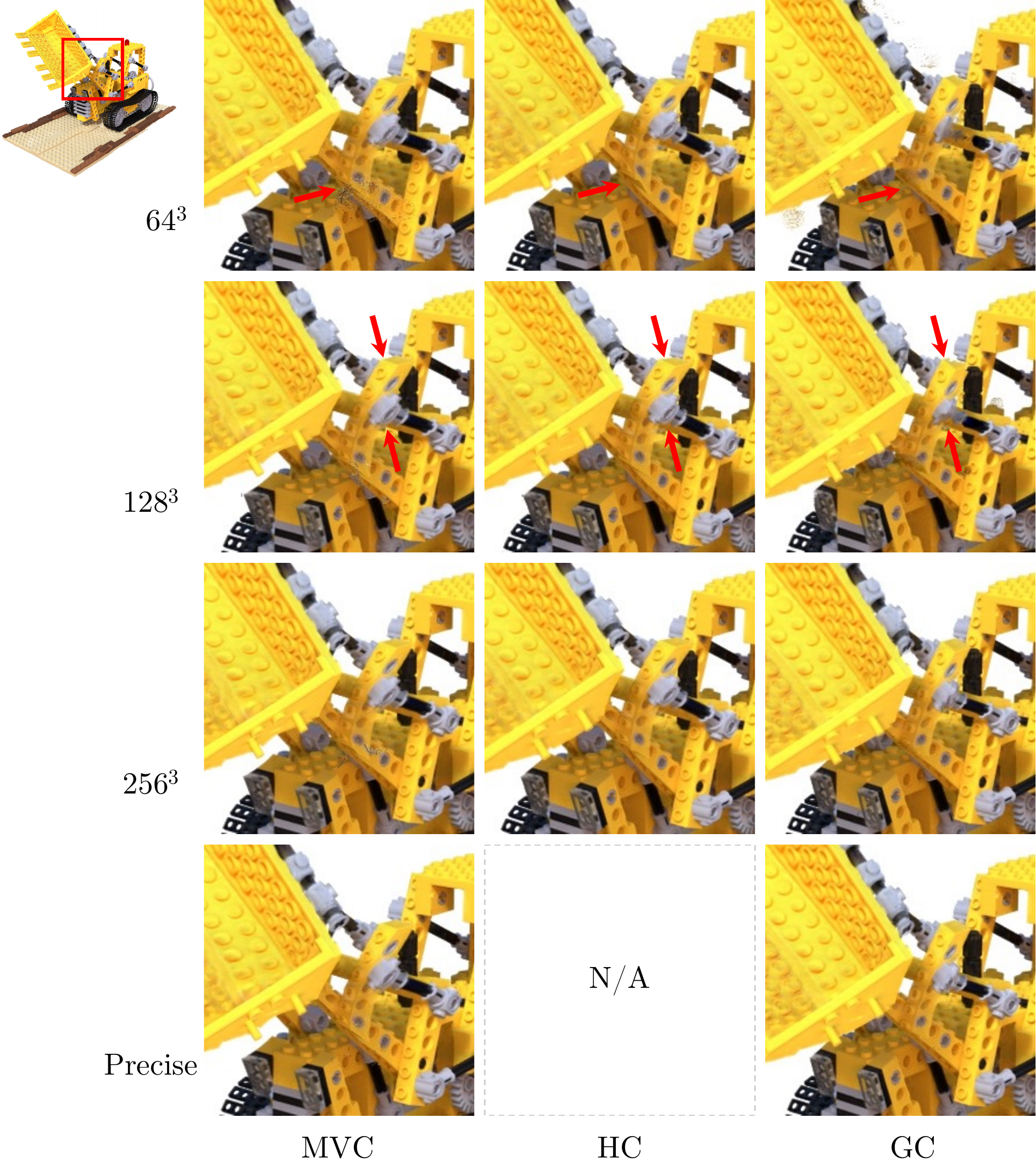}
\caption{Ablation on different cage coordinates and discretization resolution on synthetic ``Lego'' dataset. For more details please refer to Sec.~\ref{sec:ablation} and Tab.~\ref{table:abl}.}
\label{fig:abl_syn}
\end{figure}

\begin{figure}[t]
\centering
\includegraphics[width=\linewidth]{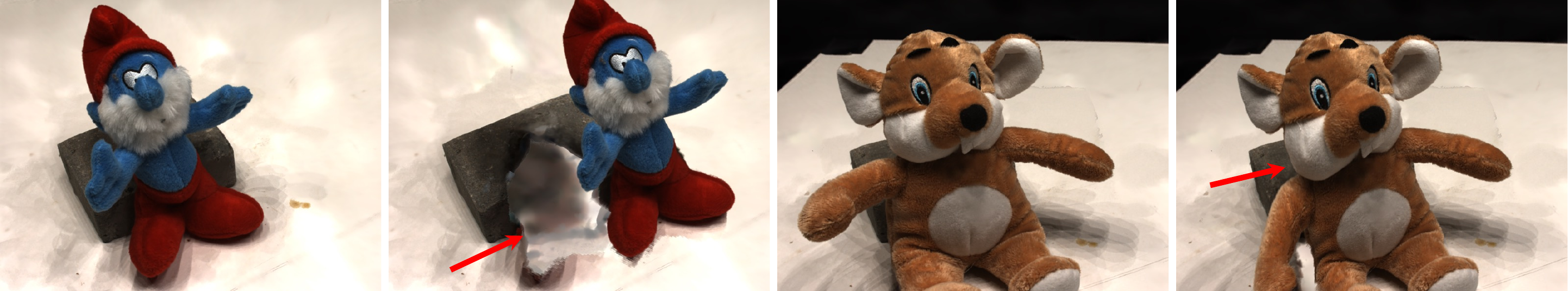}
\caption{Failure cases. Left: moving the foreground object results in exposing the parts of the scene that are under-modeled due to occlusion. Right: drastic cage manipulation may cause the artifacts.}
\label{fig:failure}
\end{figure}

\subsection{Ablation study} \label{sec:ablation}

We discuss the impact of different discretization resolutions and different cage coordinates on the synthesis quality. As introduced in~\ref{sec:related}, we use three commonly used cage coordinates for comparison: mean value coordinates (MVC)~\cite{ju2005mean}, harmonic coordinates~\cite{joshi2007harmonic} and green coordinates~\cite{lipman2008green}. We observed that the impact of the above two factors on the synthesis quality is subtle, we choose the synthetic ``Lego'' scene with relatively obvious distinction for ablation. The computation time for rendering an image and the synthesis results are shown in Tab.~\ref{table:abl} and Fig~\ref{fig:abl_syn}, respectively. 

We use the same settings as assumptions in Sec.~\ref{sec:fast} except for the number of rendered images, i.e. $h=w=800, M=512, n=128$ and $N=1$. The cage we used for the synthetic ``Lego'' scene has 42 vertices.

\paragraph{\textbf{Impact of discretization resolution.}}
As shown in Tab.~\ref{table:abl} and Fig~\ref{fig:abl_syn}, although $64^3$ resolution has a faster computation speed, significant artifacts can be observed in the synthesized results (e.g. blurry or fake shadow). This indicates that low discretization resolution brings a large error in the deformed-to-canonical mapping of the sampled points. For $128^3$ resolution, the artifact is lightened with an acceptable computation time increase. For $256^3$ resolutions, it can be seen that the improvement in synthesized quality is limited, but causes a larger increase in computation time as well as memory cost. For cage coordinates that have a closed-formulation (i.e., MVC and GC), although it is impractical due to the extremely long computation time (about $2\sim4$ minutes per scene), we show the results of the precise computation of cage coordinates without using discretization as an upper limit for comparison.

\paragraph{\textbf{Impact of different cage coordinates.}}
Comparison are also shown in Fig~\ref{fig:abl_syn}. MVC and HC show similar synthesized quality. GC shows a more reasonable deformation for the long-striped parts in the center of the image, however, the loss of detail for small parts is also observed. 

\subsection{Limitations}

\paragraph{\textbf{Cage generation.}}
As noticed in the cage-based mesh deformation, the quality of the cage greatly affects the deformation quality. However, the method we use for cage generation shows some difficulties in representing detailed cage shapes while keeping a small number of cage vertices. For objects with difficult shapes, manual refinement of the cage comes necessary, especially for real scenes with backgrounds. We conducted an extensive survey on the automatic cage generation from 3D scenes, and to our surprise, this task seems to still be unexplored. Especially for real scenes, to the best of our knowledge, there is no effective way to generate a cage automatically. We believe that the automatic generation of the cage from 3D scenes is a promising direction for future work.

\paragraph{\textbf{Failure cases.}}
We report some failure cases of our approach in Fig.~\ref{fig:failure}. The first typical failure case occurs when a part of the scene (usually the background) is not well modeled due to the occlusion. When deforming or moving the foreground objects so that the under-modeled part is exposed, obvious artifacts will be observed. However, addressing this issue is very challenging because the traditional optimization method of the radiance field cannot handle the part unseen during the training, which makes other scene manipulation methods~\cite{yang2021objectnerf} also suffer from the same issue. We assume that the use of occlusion-aware scene modeling methods or scene completion techniques may help to alleviate this issue. 

The second typical failure case may occur when a part of the object gets drastically deformed, the irrelevant parts may also be affected thus causing artifacts. This is also a long-standing issue for cage-based mesh deformation. As discussed in the previous works of CBD~\cite{nieto2013cage}, we believe that this issue might be alleviated by generating cages with higher accuracy as mentioned above or by choosing appropriate cage coordinates.

\section{Conclusion}
We presented a new method that enables free-form deformation of the radiance field. We derived a novel formulation to extend the application of cage-based deformation to the radiance field. By manipulating the vertices of the cage, we can explicitly perform free-form deformation of the radiance field while maintaining photorealistic rendering quality. To address the issue of impractical deformation computation time that arises in a naive implementation, we propose to use a discretization method specifically adapted for the radiance field and succeed in reducing the computation time by several orders of magnitude. Currently, the quality of the scene deformation is still largely influenced by the quality of the generated cage, this leaves us with a trade-off between the effort of manual cage refinement and the deformation quality. A better automatic cage generation algorithm would be a promising direction for future work.

\subsubsection{Acknowledgements} We would like to thank Daisuke Kasuga, Ryosuke Sasaki, Tomoyuki Takahata, Haruo Fujiwara, and Atsuhiro Noguchi for comments and discussions. This work was partially supported by JST AIP Acceleration Research JPMJCR20U3, Moonshot R\&D Grant Number JPMJPS2011, CREST Grant Number JPMJCR2015, JSPS KAKENHI Grant Number JP19H01115 and Basic Research Grant (Super AI) of Institute for AI and Beyond of the University of Tokyo.

\clearpage
\bibliographystyle{splncs04}
\bibliography{6719}

\begin{thebibliography}{10}
\providecommand{\url}[1]{\texttt{#1}}
\providecommand{\urlprefix}{URL }
\providecommand{\doi}[1]{https://doi.org/#1}

\bibitem{barron2021mip}
Barron, J.T., Mildenhall, B., Tancik, M., Hedman, P., Martin-Brualla, R.,
  Srinivasan, P.P.: Mip-nerf: A multiscale representation for anti-aliasing
  neural radiance fields. In: Proceedings of the IEEE/CVF International
  Conference on Computer Vision. pp. 5855--5864 (2021)

\bibitem{tensorf}
Chen, A., Xu, Z., Geiger, A., Yu, J., Su, H.: Tensorf: Tensorial radiance
  fields. European Conference on Computer Vision  (2022)

\bibitem{blender}
Community, B.O.: Blender - a 3D modelling and rendering package. Blender
  Foundation, Stichting Blender Foundation, Amsterdam (2018),
  \url{http://www.blender.org}

\bibitem{davis2012unstructured}
Davis, A., Levoy, M., Durand, F.: Unstructured light fields. In: Computer
  Graphics Forum. vol.~31, pp. 305--314. Wiley Online Library (2012)

\bibitem{derose2006harmonic}
DeRose, T., Meyer, M.: Harmonic coordinates. In: Pixar Technical Memo 06-02,
  Pixar Animation Studios (2006)

\bibitem{fan2017point}
Fan, H., Su, H., Guibas, L.J.: A point set generation network for 3d object
  reconstruction from a single image. In: Proceedings of the IEEE conference on
  computer vision and pattern recognition. pp. 605--613 (2017)

\bibitem{floater2003mean}
Floater, M.S.: Mean value coordinates. Computer aided geometric design  (2003)

\bibitem{yu_and_fridovichkeil2021plenoxels}
Fridovich-Keil, S., Yu, A., Tancik, M., Chen, Q., Recht, B., Kanazawa, A.:
  Plenoxels: Radiance fields without neural networks. In: Proceedings of the
  IEEE/CVF Conference on Computer Vision and Pattern Recognition. pp.
  5501--5510 (2022)

\bibitem{girdhar2016learning}
Girdhar, R., Fouhey, D.F., Rodriguez, M., Gupta, A.: Learning a predictable and
  generative vector representation for objects. In: European Conference on
  Computer Vision. pp. 484--499. Springer (2016)

\bibitem{gortler1996lumigraph}
Gortler, S.J., Grzeszczuk, R., Szeliski, R., Cohen, M.F.: The lumigraph. In:
  Proceedings of the 23rd annual conference on Computer graphics and
  interactive techniques. pp. 43--54 (1996)

\bibitem{guo2021pct}
Guo, M.H., Cai, J.X., Liu, Z.N., Mu, T.J., Martin, R.R., Hu, S.M.: Pct: Point
  cloud transformer. Computational Visual Media  (2021)

\bibitem{guo2020osf}
Guo, M., Fathi, A., Wu, J., Funkhouser, T.: Object-centric neural scene
  rendering. arXiv preprint arXiv:2012.08503  (2020)

\bibitem{hartley2003multiple}
Hartley, R., Zisserman, A.: Multiple view geometry in computer vision.
  Cambridge university press (2003)

\bibitem{jakab2021keypointdeformer}
Jakab, T., Tucker, R., Makadia, A., Wu, J., Snavely, N., Kanazawa, A.:
  Keypointdeformer: Unsupervised 3d keypoint discovery for shape control. In:
  Proceedings of the IEEE/CVF Conference on Computer Vision and Pattern
  Recognition. pp. 12783--12792 (2021)

\bibitem{jensen2014large}
Jensen, R., Dahl, A., Vogiatzis, G., Tola, E., Aan{\ae}s, H.: Large scale
  multi-view stereopsis evaluation. In: Proceedings of the IEEE conference on
  computer vision and pattern recognition. pp. 406--413 (2014)

\bibitem{zhang2021stnerf}
Jiakai, Z., Xinhang, L., Xinyi, Y., Fuqiang, Z., Yanshun, Z., Minye, W.,
  Yingliang, Z., Lan, X., Jingyi, Y.: Editable free-viewpoint video using a
  layered neural representation. In: ACM SIGGRAPH (2021)

\bibitem{joshi2007harmonic}
Joshi, P., Meyer, M., DeRose, T., Green, B., Sanocki, T.: Harmonic coordinates
  for character articulation. ACM Transactions on Graphics (TOG)  (2007)

\bibitem{ju2005mean}
Ju, T., Schaefer, S., Warren, J.: Mean value coordinates for closed triangular
  meshes. In: ACM Siggraph 2005 Papers, pp. 561--566 (2005)

\bibitem{kajiya1984ray}
Kajiya, J.T., Von~Herzen, B.P.: Ray tracing volume densities. ACM SIGGRAPH
  computer graphics  (1984)

\bibitem{kanazawa2018learning}
Kanazawa, A., Tulsiani, S., Efros, A.A., Malik, J.: Learning category-specific
  mesh reconstruction from image collections. In: European Conference on
  Computer Vision. pp. 371--386 (2018)

\bibitem{kato2018neural}
Kato, H., Ushiku, Y., Harada, T.: Neural 3d mesh renderer. In: Proceedings of
  the IEEE conference on computer vision and pattern recognition. pp.
  3907--3916 (2018)

\bibitem{lipman2008green}
Lipman, Y., Levin, D., Cohen-Or, D.: Green coordinates. ACM Transactions on
  Graphics (TOG)  (2008)

\bibitem{liu2020neural}
Liu, L., Gu, J., Zaw~Lin, K., Chua, T.S., Theobalt, C.: Neural sparse voxel
  fields. Advances in Neural Information Processing Systems  \textbf{33},
  15651--15663 (2020)

\bibitem{liu2021neural}
Liu, L., Habermann, M., Rudnev, V., Sarkar, K., Gu, J., Theobalt, C.: Neural
  actor: Neural free-view synthesis of human actors with pose control. ACM
  Trans. Graph.(ACM SIGGRAPH Asia)  (2021)

\bibitem{liu2021editing}
Liu, S., Zhang, X., Zhang, Z., Zhang, R., Zhu, J.Y., Russell, B.: Editing
  conditional radiance fields. In: Proceedings of the IEEE/CVF International
  Conference on Computer Vision. pp. 5773--5783 (2021)

\bibitem{lorensen1987marching}
Lorensen, W.E., Cline, H.E.: Marching cubes: A high resolution 3d surface
  construction algorithm. ACM siggraph computer graphics  (1987)

\bibitem{mildenhall2022nerf}
Mildenhall, B., Hedman, P., Martin-Brualla, R., Srinivasan, P.P., Barron, J.T.:
  Nerf in the dark: High dynamic range view synthesis from noisy raw images.
  In: Proceedings of the IEEE/CVF Conference on Computer Vision and Pattern
  Recognition. pp. 16190--16199 (2022)

\bibitem{mildenhall2020nerf}
Mildenhall, B., Srinivasan, P.P., Tancik, M., Barron, J.T., Ramamoorthi, R.,
  Ng, R.: Nerf: Representing scenes as neural radiance fields for view
  synthesis. In: European conference on computer vision. pp. 405--421. Springer
  (2020)

\bibitem{mueller2022instant}
M\"uller, T., Evans, A., Schied, C., Keller, A.: Instant neural graphics
  primitives with a multiresolution hash encoding. ACM Trans. Graph.  (2022)

\bibitem{niemeyer2020differentiable}
Niemeyer, M., Mescheder, L., Oechsle, M., Geiger, A.: Differentiable volumetric
  rendering: Learning implicit 3d representations without 3d supervision. In:
  Proceedings of the IEEE/CVF Conference on Computer Vision and Pattern
  Recognition. pp. 3504--3515 (2020)

\bibitem{nieto2013cage}
Nieto, J.R., Sus{\'\i}n, A.: Cage based deformations: a survey. In: Deformation
  models, pp. 75--99. Springer (2013)

\bibitem{noguchi2022watch}
Noguchi, A., Iqbal, U., Tremblay, J., Harada, T., Gallo, O.: Watch it move:
  Unsupervised discovery of 3d joints for re-posing of articulated objects. In:
  Proceedings of the IEEE/CVF Conference on Computer Vision and Pattern
  Recognition. pp. 3677--3687 (2022)

\bibitem{2021narf}
Noguchi, A., Sun, X., Lin, S., Harada, T.: Neural articulated radiance field.
  In: Proceedings of the IEEE/CVF International Conference on Computer Vision.
  pp. 5762--5772 (2021)

\bibitem{Ost_2021_CVPR}
Ost, J., Mannan, F., Thuerey, N., Knodt, J., Heide, F.: Neural scene graphs for
  dynamic scenes. In: Proceedings of the IEEE/CVF Conference on Computer Vision
  and Pattern Recognition. pp. 2856--2865 (2021)

\bibitem{park2021nerfies}
Park, K., Sinha, U., Barron, J.T., Bouaziz, S., Goldman, D.B., Seitz, S.M.,
  Martin-Brualla, R.: Nerfies: Deformable neural radiance fields. In:
  Proceedings of the IEEE/CVF International Conference on Computer Vision. pp.
  5865--5874 (2021)

\bibitem{park2021hypernerf}
Park, K., Sinha, U., Hedman, P., Barron, J.T., Bouaziz, S., Goldman, D.B.,
  Martin-Brualla, R., Seitz, S.M.: Hypernerf: A higher-dimensional
  representation for topologically varying neural radiance fields. ACM Trans.
  Graph.  (2021)

\bibitem{peng2021animatable}
Peng, S., Dong, J., Wang, Q., Zhang, S., Shuai, Q., Zhou, X., Bao, H.:
  Animatable neural radiance fields for modeling dynamic human bodies. In:
  Proceedings of the IEEE/CVF International Conference on Computer Vision. pp.
  14314--14323 (2021)

\bibitem{peng2021neural}
Peng, S., Zhang, Y., Xu, Y., Wang, Q., Shuai, Q., Bao, H., Zhou, X.: Neural
  body: Implicit neural representations with structured latent codes for novel
  view synthesis of dynamic humans. In: Proceedings of the IEEE/CVF Conference
  on Computer Vision and Pattern Recognition. pp. 9054--9063 (2021)

\bibitem{pumarola2021d}
Pumarola, A., Corona, E., Pons-Moll, G., Moreno-Noguer, F.: D-nerf: Neural
  radiance fields for dynamic scenes. In: Proceedings of the IEEE/CVF
  Conference on Computer Vision and Pattern Recognition. pp. 10318--10327
  (2021)

\bibitem{sitzmann2019scene}
Sitzmann, V., Zollh{\"o}fer, M., Wetzstein, G.: Scene representation networks:
  Continuous 3d-structure-aware neural scene representations. Advances in
  Neural Information Processing Systems  \textbf{32} (2019)

\bibitem{su2021anerf}
Su, S.Y., Yu, F., Zollh{\"o}fer, M., Rhodin, H.: A-nerf: Articulated neural
  radiance fields for learning human shape, appearance, and pose. Advances in
  Neural Information Processing Systems  \textbf{34},  12278--12291 (2021)

\bibitem{SunSC22}
Sun, C., Sun, M., Chen, H.T.: Direct voxel grid optimization: Super-fast
  convergence for radiance fields reconstruction. In: Proceedings of the
  IEEE/CVF Conference on Computer Vision and Pattern Recognition. pp.
  5459--5469 (2022)

\bibitem{tretschk2021non}
Tretschk, E., Tewari, A., Golyanik, V., Zollh{\"o}fer, M., Lassner, C.,
  Theobalt, C.: Non-rigid neural radiance fields: Reconstruction and novel view
  synthesis of a dynamic scene from monocular video. In: Proceedings of the
  IEEE/CVF International Conference on Computer Vision. pp. 12959--12970 (2021)

\bibitem{triggs1999bundle}
Triggs, B., McLauchlan, P.F., Hartley, R.I., Fitzgibbon, A.W.: Bundle
  adjustment—a modern synthesis. In: International workshop on vision
  algorithms. pp. 298--372. Springer (1999)

\bibitem{xian2009automatic}
Xian, C., Lin, H., Gao, S.: Automatic generation of coarse bounding cages from
  dense meshes. In: IEEE International Conference on Shape Modeling and
  Applications (2009)

\bibitem{xu2022surface}
Xu, T., Fujita, Y., Matsumoto, E.: Surface-aligned neural radiance fields for
  controllable 3d human synthesis. In: Proceedings of the IEEE/CVF Conference
  on Computer Vision and Pattern Recognition. pp. 15883--15892 (2022)

\bibitem{yan2016perspective}
Yan, X., Yang, J., Yumer, E., Guo, Y., Lee, H.: Perspective transformer nets:
  Learning single-view 3d object reconstruction without 3d supervision.
  Advances in neural information processing systems  \textbf{29} (2016)

\bibitem{yang2021objectnerf}
Yang, B., Zhang, Y., Xu, Y., Li, Y., Zhou, H., Bao, H., Zhang, G., Cui, Z.:
  Learning object-compositional neural radiance field for editable scene
  rendering. In: Proceedings of the IEEE/CVF International Conference on
  Computer Vision. pp. 13779--13788 (2021)

\bibitem{Yifan:NeuralCage:2020}
Yifan, W., Aigerman, N., Kim, V.G., Chaudhuri, S., Sorkine-Hornung, O.: Neural
  cages for detail-preserving 3d deformations. In: Proceedings of the IEEE/CVF
  Conference on Computer Vision and Pattern Recognition. pp. 75--83 (2020)

\bibitem{yuan2022nerf-editing}
Yuan, Y.J., Sun, Y.T., Lai, Y.K., Ma, Y., Jia, R., Gao, L.: Nerf-editing:
  geometry editing of neural radiance fields. In: Proceedings of the IEEE/CVF
  Conference on Computer Vision and Pattern Recognition. pp. 18353--18364
  (2022)

\end{thebibliography}
\end{document}